\documentclass[letterpaper, 10 pt, conference]{ieeeconf}  

\IEEEoverridecommandlockouts
\usepackage{tabularx}
\usepackage{caption}

\usepackage{subcaption}
\usepackage[utf8]{inputenc}
\usepackage[T1]{fontenc}
\usepackage{xcolor}
\definecolor{citecolor}{HTML}{0071bc}
\definecolor{shadecolor}{rgb}{0.94,0.94,0.94}

\usepackage[hidelinks]{hyperref}
\usepackage[normalem]{ulem}
\usepackage{url}
\usepackage{booktabs}
\usepackage{amsfonts}
\usepackage{amsmath}
\usepackage{microtype}
\usepackage{graphicx}

\newcommand\mypara[1]{\vspace{0mm}\noindent\textbf{#1}}

\title{\LARGE\bfseries
Data-Driven Risk Fields for Safer End-to-End Autonomous Driving
}

\author{Yuanxin Tian$^{1,\dagger}$, Zhiyuan Liu$^{1,\dagger}$, Jinhao Li$^{1}$, Liangfan Zhu$^{2}$, Shuai Wang$^{2}$, Heye Huang$^{3}$\\
Qingwen Meng$^{1}$, Fang Zhang$^{1}$, Liuzhu Tong$^{2}$, Zhenhua Xu$^{1,*}$, Wenhao Yu$^{1,*}$, and Jianqiang Wang$^{1}$\\
{\normalfont\footnotesize $^{1}$Tsinghua University \quad $^{2}$Yinwang \quad $^{3}$KAIST}%
\thanks{\scriptsize $^{\dagger}$Equal contribution; $^{*}$Co-corresponding authors: Zhenhua Xu and Wenhao Yu.}}

\begin{document}
\bstctlcite{IEEEtran:BSTcontrol}

\maketitle
\thispagestyle{empty}
\pagestyle{empty}

\begin{figure*}[t]
\vspace{-1em}
    \centering
    \includegraphics[width=\textwidth]{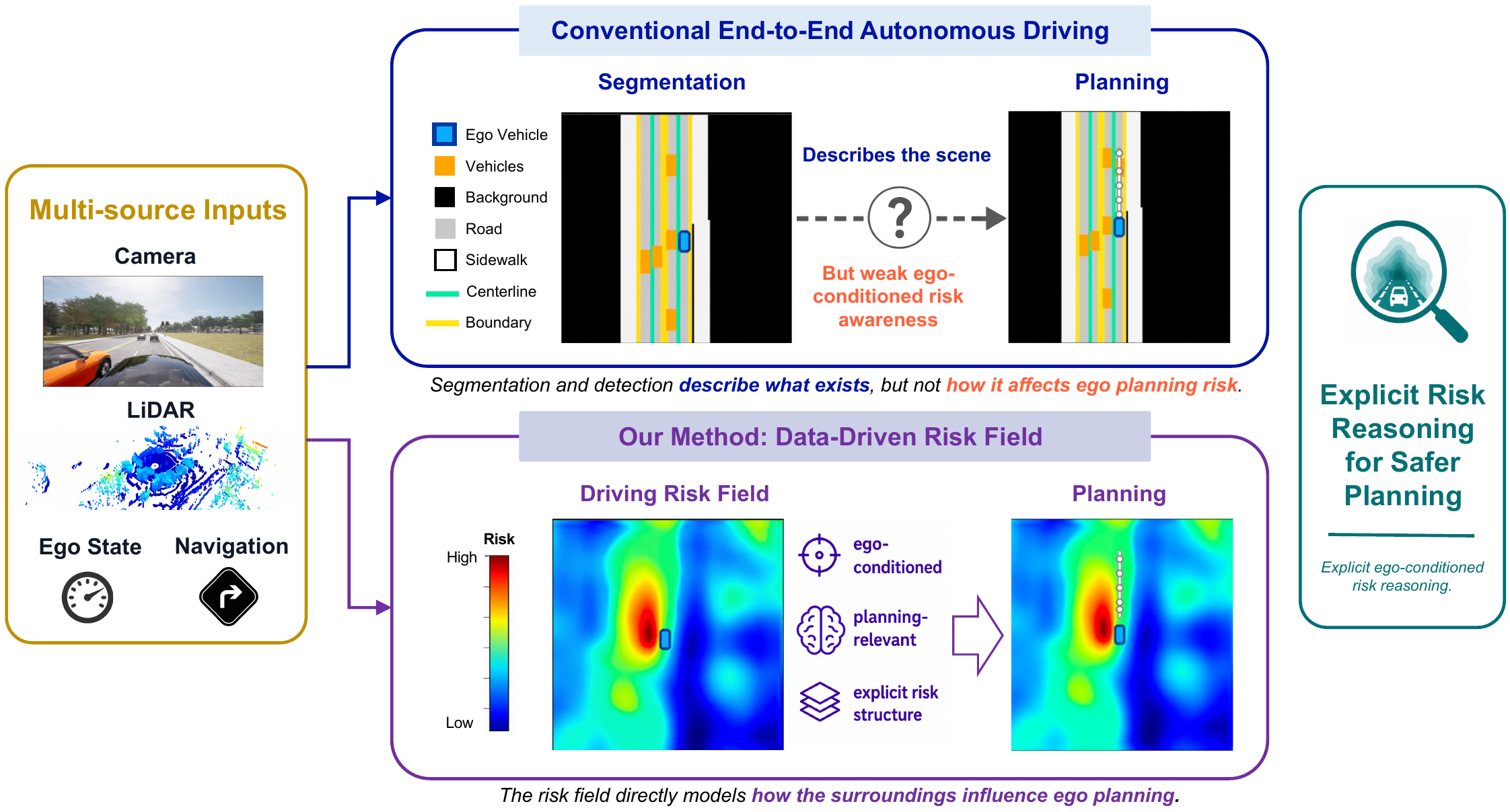}
    \caption{
    \textbf{Explicit risk modeling beyond scene understanding.}
Conventional scene representations describe objects, lanes, and semantic regions, but do not explicitly reveal how they affect ego planning.
Our method learns an ego-conditioned data-driven risk field that models how surrounding agents influence the ego vehicle, enabling safer end-to-end planning.
}
    \label{fig:teaser}
\end{figure*}

\begin{abstract}
Safety is a fundamental requirement for autonomous driving, yet existing end-to-end driving models still lack explicit risk-aware learning capacities. Existing rule-based risk models provide interpretable safety priors, yet their absolute risk scores depend on handcrafted functions, coefficients, and thresholds. Learning-based risk representations reduce part of this manual design, but their supervision often relies on occupancy-derived labels or heuristic cost values, which may not capture ego-conditioned planning risk. In this paper, we propose \textbf{DRiF}, a data-driven risk-field framework for safer end-to-end autonomous driving. DRiF learns a shared BEV feature with static map segmentation, dynamic risk prediction, and vehicle planning. For dynamic risk learning, DRiF converts rule-based safety priors into pairwise risk labels, and trains the risk field to preserve relative risk ordering instead of regressing handcrafted absolute scores. Experiments on Bench2Drive show that DRiF achieves competitive overall performance, with consistent improvements in driving score, success rate, and collision-related metrics. These results establish relative risk supervision as an effective way to connect explicit safety structure with end-to-end planning. The data and code will be publicly available.
\end{abstract}

\section{Introduction}

Autonomous driving has progressed rapidly in both academia and industry, with data-driven end-to-end driving becoming a representative paradigm. Given sensory observations and route commands, an end-to-end model directly predicts future trajectories \cite{bojarski2016end,bansal2019chauffeurnet,chitta2023transfuser,chen2024end}. Early methods established the feasibility of supervised sensor-to-action learning. Subsequent systems strengthen this formulation by introducing structured scene modeling into the learning pipeline. BEV encoders provide a unified top-down space for multi-view perception \cite{philion2020lift,li2022bevformer,liu2022bevfusion}; online mapping and vectorized scene modeling recover road geometry and topology \cite{liao2022maptr,jiang2023vad,xu2024insmapper}; motion prediction further extends the representation from static layout to future scene evolution \cite{hu2021fiery,shi2022motion,wei2023surroundocc,zhang2023occformer}. Finally, planning-oriented architectures connect these intermediate outputs with ego trajectory generation \cite{hu2022stp3,hu2023uniad,sun2024sparsedrive}. Recent planners further improve trajectory modeling through generative policies, temporal-consistency modeling, and driving world models \cite{liao2025diffusiondrive,xing2025goalflow,song2025momad,li2025wote,zheng2025world4drive,zhang2025seerdrive,zhang2026resworld}. Taken together, these advances make end-to-end driving an increasingly powerful paradigm. 

Safety is the central requirement of autonomous driving. A reliable vehicle must not only perceive the scene, but also understand how surrounding objects, road structures, predicted motions, and route constraints affect its own future behavior. Existing end-to-end planners usually learn this perception-to-planning relation implicitly through latent features and final trajectory losses \cite{hu2023uniad,jiang2023vad,sun2024sparsedrive,liao2025diffusiondrive}. Such supervision can capture common expert driving patterns, but it does not explicitly indicate which BEV regions are dangerous to the current ego vehicle, why they are dangerous, or how their risk should influence planning. Without explicit risk-oriented supervision, an end-to-end planner may fail to learn reliable safety structure, especially in long-tail scenarios such as cut-in, merging, intersections, pedestrian crossing, lane-boundary violation, and off-road behavior.

A direct way to introduce safety structure is to model driving risk explicitly. Classical traffic safety metrics assign criticality to interactions from different perspectives. Time-to-collision estimates collision imminence under continued motion \cite{hayward1972near}; time headway measures longitudinal following safety \cite{vogel2003comparison}; post-encroachment time evaluates temporal separation at a conflict point \cite{cooper1984experience}; deceleration rate to avoid collision measures the braking effort required to prevent a crash \cite{cooper1976traffic,fu2021comparison}. Beyond scalar metrics, risk-field methods provide a more general way to represent safety as spatial cost. Artificial potential fields first model obstacles as repulsive forces \cite{khatib1986real}, while probabilistic risk fields further incorporate motion uncertainty and collision probability \cite{mullakkal2020probabilistic,wang2022probabilistic,wang2026data}. Among these methods, driving safety fields offer a particularly relevant formulation for autonomous driving, as they encode driver--vehicle--road interactions in a unified continuous field and provide an interpretable spatial representation for trajectory evaluation and safety-aware planning \cite{wang2015driving,wang2016driving}. Although these formulations are interpretable and auditable, their risk magnitudes usually depend on hand-designed coefficients, thresholds, and scenario-specific calibration. This manual design limits their scalability and makes them difficult to use as direct supervision for end-to-end representation learning.

Learning-based risk representations reduce part of this manual design, but their risk modeling and supervision remain limited. RiskMap learns a differentiable risk field as a driving cost prior for motion planning \cite{xin2024riskmap}; Risk Occupancy extends occupancy-style modeling with spatial, temporal, and risk dimensions under vehicle-road-cloud collaboration \cite{chen2024riskoccupancy}; RiskMM uses risk maps as middleware for cooperative end-to-end planning \cite{lei2025riskmm}. These methods bring risk modeling closer to data-driven planning systems. However, driving risk does not have an objective ground truth comparable to depth, occupancy, or object location. Existing models therefore often rely on occupancy-derived labels or heuristic planner scores. Such supervision may simplify risk into occupied-or-costly regions, without explicitly deriving how the surrounding agents jointly determine planning risk. As a result, the learned risk map may remain an occupancy-like or cost-map-like surrogate, rather than a genuine ego-conditioned risk representation. This can bind the network to a particular handcrafted risk scale and limit its ability to identify and avoid risks in complex or long-tail driving scenarios.

We address the aforementioned problems by decomposing BEV driving risk into a static map and a dynamic interactive risk field. The static map captures hard constraints from route feasibility and ego reachability, which is supervised with dense labels. The dynamic interactive risk field models planning risk induced by surrounding-agent futures and potential conflicts, where calibrated dense labels are unavailable. We therefore learn it from a relative ranking loss between sampled BEV locations. Based on this design, we propose \textbf{DRiF}, an end-to-end autonomous driving framework relying on \textbf{D}ata-driven \textbf{Ri}sk-\textbf{F}ield. DRiF encodes multi-source driving inputs into a shared BEV representation and predicts both static map and dynamic interactive risk. During training, a pairwise risk label generator converts rule-based safety priors into pairwise ranking labels to supervise the dynamic risk head, thereby injecting explicit risk structure into the representation used by the end-to-end planner.

Our contributions are summarized as follows:
\begin{itemize}
    \item We propose \textbf{DRiF}, an end-to-end autonomous driving framework that learns ego-conditioned risk representations in a data-driven manner for trajectory planning. 
    \item We introduce relative-risk supervision, which learns driving risk through pairwise comparisons instead of hand-crafted absolute scalar labels.
    \item We conduct extensive experiments on the large closed-loop autonomous driving benchmark Bench2Drive. DRiF achieves competitive performance across overall, safety, and multi-ability metrics, demonstrating the effectiveness of explicit risk modeling for safer end-to-end planning.
\end{itemize}

\section{Related Work}

\mypara{End-to-End Autonomous Driving.} End-to-end autonomous driving learns ego actions from onboard observations and expert demonstrations, with outputs ranging from control commands to future trajectories \cite{bojarski2016end,bansal2019chauffeurnet,chitta2023transfuser,chen2024end,xu2024drivegpt4,xu2025drivegpt4v2}. Recent systems improve this formulation by introducing stronger scene representations. BEV-based methods build a unified spatial feature space for multi-view perception \cite{philion2020lift,li2022bevformer,liu2022bevfusion}; online mapping and vectorized scene modeling encode road structures for downstream reasoning \cite{liao2022maptr,jiang2023vad}; motion and occupancy prediction further enable the planner to reason about future scene evolution \cite{hu2021fiery,shi2022motion,wei2023surroundocc,zhang2023occformer}. Representative planning-oriented frameworks include UniAD, which jointly models perception, prediction, occupancy, and planning \cite{hu2023uniad}; VAD, which uses vectorized scene elements for efficient trajectory generation \cite{jiang2023vad}; and SparseDrive, which formulates end-to-end planning with sparse scene representation \cite{sun2024sparsedrive}. More recent methods strengthen trajectory modeling through generative policies, temporal consistency, and world models \cite{liao2025diffusiondrive,xing2025goalflow,song2025momad,li2025wote}. Safety-oriented planners further improve interaction handling through path-conditioned longitudinal planning and safety-critical augmentation \cite{wu2026aligndrive}, ego-centric joint-causal modeling and policy alignment \cite{moon2026caad}, or future-scene priors from 4D occupancy world models \cite{cheng2026owmdrive}. These works improve planning structure, policy alignment, or future-scene modeling, whereas ego-conditioned risk supervision in a shared representation remains under-explored.

\mypara{Risk Representations for Driving.} Classical safety metrics quantify interaction criticality using time-to-collision, time headway, post-encroachment time, deceleration rate, or formal safe-distance constraints \cite{hayward1972near,vogel2003comparison,cooper1984experience,cooper1976traffic,fu2021comparison,shalev2017formal}. Risk-field methods represent safety as spatial cost through artificial potential fields, probabilistic risk fields, or continuous driver--vehicle--road interaction fields \cite{khatib1986real,mullakkal2020probabilistic,wang2022probabilistic,wang2026data,wang2015driving,wang2016driving}. Recent learning-based approaches establish complementary risk-to-planning interfaces: RiskMap learns a differentiable risk field \cite{xin2024riskmap}; Risk Occupancy extends occupancy representations with risk dimensions \cite{chen2024riskoccupancy}; RiskMM learns an interpretable spatiotemporal risk representation and feeds it to learning-based MPC \cite{lei2025riskmm}. Although these studies demonstrate the value of risk-aware planning, their supervision often relies on manually calibrated risk functions or occupancy-derived labels, requiring models to regress uncertain absolute risk values.

\mypara{Relative Supervision.} Relative supervision is effective when absolute labels are ambiguous or difficult to calibrate. Ranking formulations learn from pairwise preferences \cite{bradley1952rank,burges2005learning}, and ordinal depth methods show that useful spatial structure can be learned without a globally calibrated scale \cite{chen2016single,fu2018deep,yang2024depthanythingv2}. These works show that relative constraints can provide robust supervision when absolute values lack a stable scale. DRiF brings this principle to autonomous driving risk representation by learning whether one BEV location is riskier or safer than another, thereby converting rule-based safety knowledge into planner-compatible supervision.

\section{Problem Definition}

\subsection{Ego-Conditioned BEV Risk Field}

At time $t$, an end-to-end driving system receives multi-source observations $o_t$, ego state $s_t$, navigation command $c_t$, and navigation point $p_t$. These inputs are encoded into a bird's-eye-view (BEV) domain $\Omega \subset \mathbb{R}^{2}$, which is discretized into an $H' \times W'$ grid $F_t$. We denote a continuous BEV location by $x=(x,y)$ and its corresponding grid index by $q=(u,v)$.

The goal of DRiF is to learn an ego-conditioned BEV risk field:
\[
R_t = \Psi(F_t) = \Psi(\Phi_{\theta}(o_t, s_t, c_t, p_t)),
\]
where $R_t$ describes how each BEV location affects the safety of the current ego vehicle. Different from occupancy or object detection, this representation is not an objective scene attribute. Its value depends on ego state, route intention, reachability, road structure, surrounding-agent futures, their interactions, etc.

\subsection{Risk Field Decomposition}

We decompose BEV driving supervision into two complementary outputs:
\[
R_t = \left(M_t, R_t^{dyn}\right),
\]
where $M_t \in \mathbb{R}^{H \times W \times K}$ and $R_t^{dyn} \in \mathbb{R}^{H \times W}$.

The static map representation $M_t$ describes dense BEV map semantics, such as road boundary, lane marking, lane centerline, etc. Since these semantics can be obtained from map annotations or BEV segmentation labels, this branch can be supervised with absolute values. Although $M_t$ is not itself a risk field, it provides structured road-geometry and constraint information that supports ego-conditioned risk learning and downstream planning.

The dynamic interactive risk field $R_t^{dyn}$ represents planning risk induced by surrounding-agent futures and potential conflicts. Unlike static map semantics, this risk does not have calibrated dense ground-truth values. For example, a future conflict region may be riskier than a free lane-center region, but the exact numerical risk gap is not objectively defined. Therefore, we define $R_t^{dyn}$ through relative ordering rather than absolute values.

\subsection{Relative-Risk Supervision}

For dynamic interactive risk, we construct pairwise supervision over planning-relevant BEV locations. Given a sampled pair $(x_i,x_j)$, its label is $y_{ij} \in \{+1,0,-1\}$, where $y_{ij}=+1$ means $x_i$ is riskier than $x_j$, $y_{ij}=-1$ means $x_i$ is safer than $x_j$, and $y_{ij}=0$ means the two locations are comparable. The desired ordering is $R_t^{dyn}(q_i) > R_t^{dyn}(q_j)$ if $y_{ij}=+1$, where $q_i$ and $q_j$ are the BEV grid indices of $x_i$ and $x_j$.

In short, DRiF uses dense supervision for static constraints and relative supervision for ambiguous dynamic risk. This avoids forcing static constraints and dynamic interactions onto the same numerical scale, while still providing explicit ego-conditioned risk supervision for end-to-end planning.

\begin{figure*}[t]
    \centering
    \includegraphics[width=\textwidth]{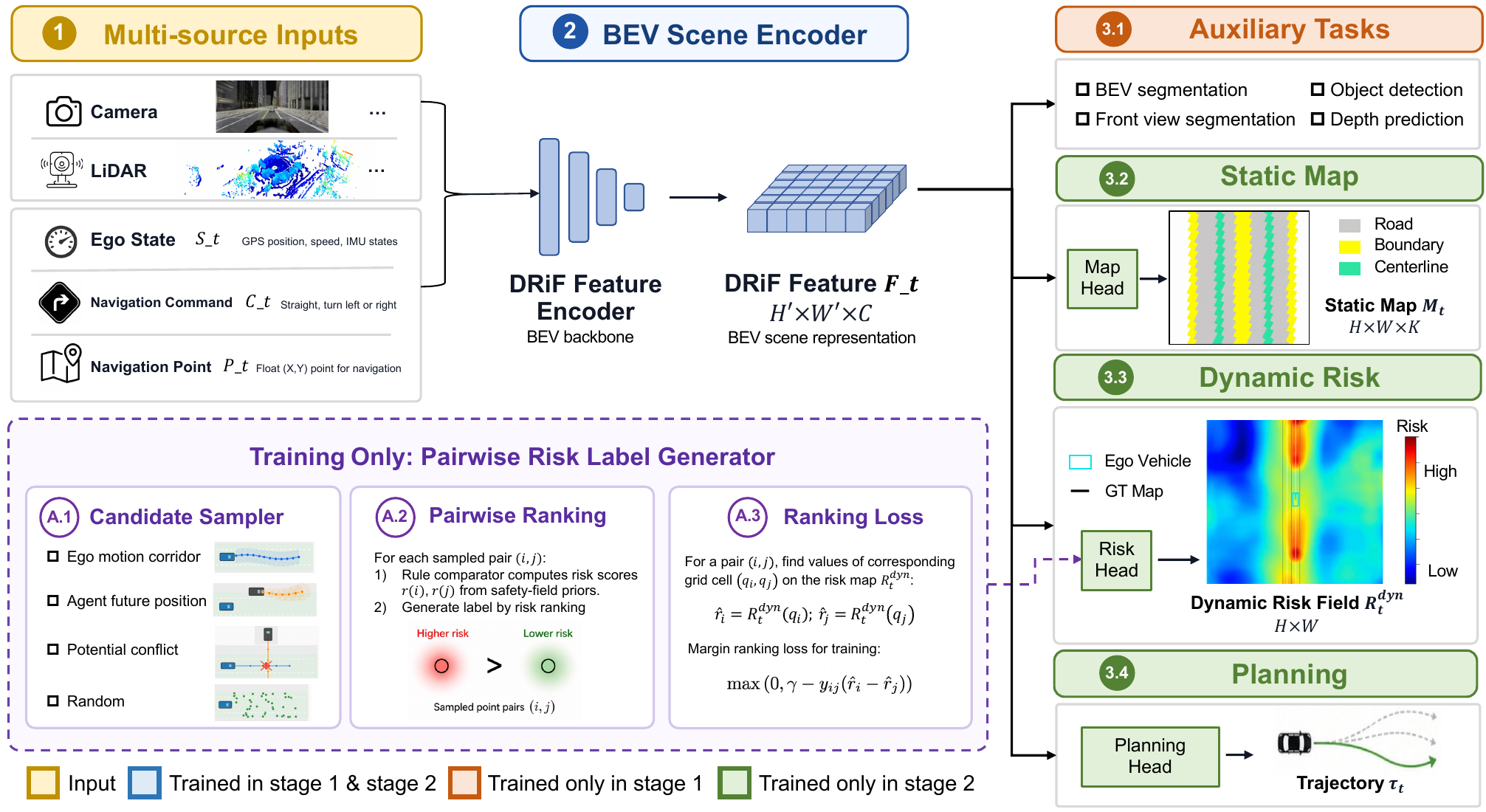}
    \caption{
    \textbf{Overview of DRiF}. 
DRiF encodes multi-source driving inputs into a shared BEV feature $F_t$ for auxiliary perception, static map segmentation, dynamic risk prediction, and trajectory planning. 
The model is trained in two stages. 
Stage 1 uses auxiliary tasks to pretrain the encoder with directly observable scene knowledge (e.g., object states). 
Stage 2 removes the auxiliary perception branch and jointly trains the static map, dynamic risk, and planning branches. The dynamic risk head is supervised by pairwise risk labels generated from rule-based safety priors, while the static map head maintains the model's understanding of static surroundings. 
    }
    \label{fig:rive_overview}
\vspace{-2em}
\end{figure*}

\section{Methodology}

\subsection{Overview}

DRiF is a risk-centric end-to-end autonomous driving framework. Given multi-source driving inputs, DRiF first extracts a shared BEV feature $F_t$. The feature is then used by four branches: an auxiliary perception branch, a static map branch, a dynamic risk branch, and a planning branch. The static map branch preserves the model's understanding of road structure, while the dynamic risk branch learns ego-conditioned interaction risk that is difficult to annotate with dense scalar labels. The overall structure of DRiF is visualized in Fig. \ref{fig:rive_overview}.

DRiF is trained in two stages. Stage 1 uses auxiliary perception tasks, such as detection, segmentation, and depth prediction, to pretrain the encoder with directly observable scene knowledge. Stage 2 removes the auxiliary branch and jointly trains static map segmentation, dynamic risk learning, and trajectory planning. During this stage, a training-only risk-pair generator converts rule-based safety priors into pairwise ranking labels for supervising the dynamic risk branch. At inference time, the label generator is removed, the learned risk representation supports planning through shared BEV features.

\subsection{Model Structure}

\mypara{BEV Encoder.}
At time $t$, the model receives observations $o_t$, ego state $s_t$, navigation command $c_t$, and navigation point $p_t$. An encoder maps them into a BEV feature $F_t=\Phi_{\theta}(o_t,s_t,c_t,p_t)$, where $F_t\in\mathbb{R}^{H'\times W'\times C}$. The encoder can be instantiated with camera, LiDAR, or multi-modal BEV backbones. Ego state and navigation information are injected so that the feature is conditioned on the current driving context.

\mypara{Auxiliary Perception Branch.}
In stage 1, the auxiliary branch predicts directly observable scene quantities, including object states, semantic segmentation, and depth. It provides strong pretraining knowledge for geometry and object-level scene understanding, and is removed during stage 2.

\mypara{Static Map Branch.}
The static map branch predicts dense BEV map semantics $\hat{M}_t=\Psi_{\mathrm{map}}(F_t)$, where $\hat{M}_t\in\mathbb{R}^{H\times W\times K}$. It is supervised by BEV map labels, such as drivable area, road boundary, lane marking, and lane centerline. This branch maintains the model's understanding of static road structures during risk learning.

\mypara{Dynamic Risk Branch.}
The dynamic risk branch predicts an ego-conditioned risk field $\hat{R}_t^{dyn}=\Psi_{\mathrm{risk}}(F_t)$, where $\hat{R}_t^{dyn}\in\mathbb{R}^{H\times W}$. Unlike occupancy or semantic maps, $\hat{R}_t^{dyn}$ represents relative planning risk caused by surrounding-agent futures and potential conflicts. Since dense calibrated risk labels are unavailable, this branch is learned with relative supervision.

\mypara{Planning Branch.}
The planning branch predicts the future ego trajectory $\hat{\tau}_t=\Psi_{\mathrm{plan}}(F_t)=\{\hat{x}_{t+k}\}_{k=1}^{T}$. It follows the original end-to-end driving objective, while benefiting from the risk-aware BEV feature shaped by static map and dynamic risk supervisions.

\subsection{Relative Supervision}
Since calibrated dense labels for dynamic risk are unavailable, DRiF learns the dynamic risk field from pairwise relative supervision. The supervision pipeline contains three steps: sampling candidate points, constructing risk pairs, and generating pairwise risk labels with rule-based safety priors. The generated labels supervise the dynamic risk branch to preserve relative risk ordering.

\mypara{Candidate Sampling.}
For each frame, DRiF first samples a set of planning-relevant BEV locations $S_t$ around the ego future trajectory, surrounding-agent futures, potential conflict regions, and route-consistent reachable areas. This step only collects candidate points that may affect ego planning.

\mypara{Pair Construction.}
After obtaining the candidate set, DRiF constructs a compact pair set $\mathcal{Q}_t=\{(x_i,x_j)\mid x_i,x_j\in S_t,i\neq j\}$. In practice, we do not enumerate all pairs. Instead, we sample pairs from different candidate sources and ego-relevant regions to avoid redundant easy comparisons and reduce computation.

\begin{figure*}[t]
    \centering
    \includegraphics[width=\textwidth]{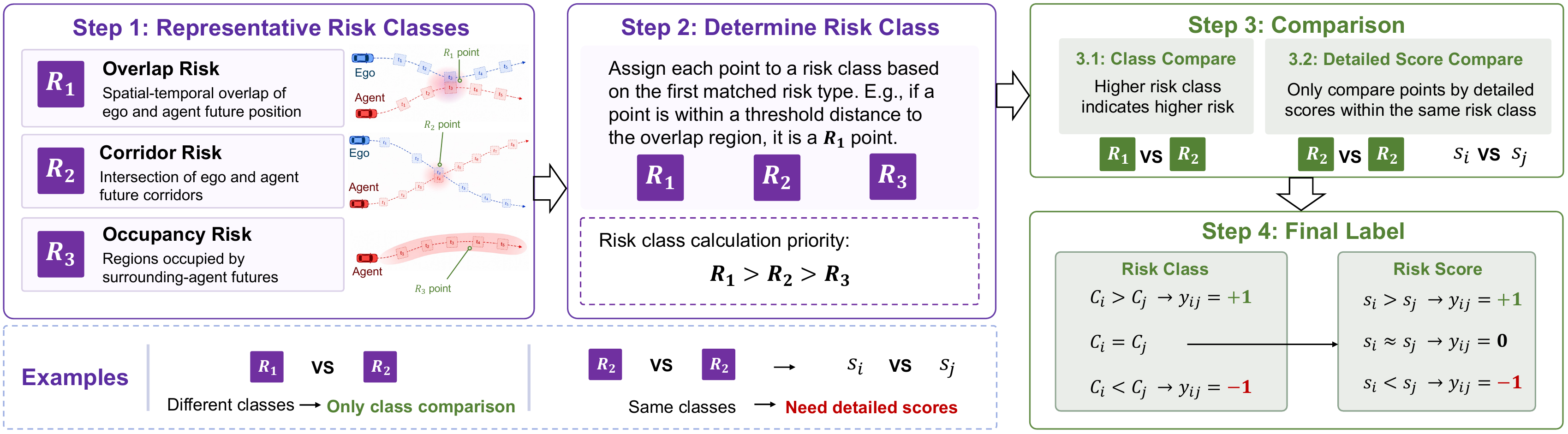}
    \caption{
    \textbf{Pairwise risk label generation}.
    Each point is first assigned to a risk class according to the matched risk type.
    The comparator then determines the pairwise risk relation by class priority; only when the two points belong to the same risk class, their detailed risk scores are further compared to generate the final label \(y_{ij}\in\{+1,0,-1\}\).
    }
    \label{fig:label_generation}
\vspace{-1.2em}
\end{figure*}

\mypara{Pairwise Risk Scoring.}
For each candidate pair $(x_i,x_j)\in\mathcal{Q}_t$, rule-based safety priors are used to evaluate their relative risk. Each point in the pair is assigned a risk signature $\phi_t(x)=[s_1(x),s_2(x),s_3(x)]$, where $s_1$ measures same-time ego-agent conflict (i.e., overlap risk), $s_2$ measures cross-time trajectory-corridor conflict (i.e., corridor risk), and $s_3$ measures surrounding-agent future occupancy (i.e., occupancy risk). These scores are not used as dense regression targets, but only as local evidence for pairwise label generation. The labeling process is visualized in Fig. \ref{fig:label_generation}.

\mypara{Pairwise Label Generation.}
For a pair $(x_i,x_j)$, DRiF first assigns each point to a risk class $C$ according to the matched risk source. We use three ordered classes, $R_1 > R_2 > R_3$, where $R_1$ denotes overlap risk, $R_2$ denotes corridor risk, and $R_3$ denotes occupancy risk. For a point $x$, the generator checks the risk scores in this priority order: if $s_1(x)$ exceeds its threshold, $x$ is assigned to $R_1$; otherwise, it checks $s_2(x)$ and then $s_3(x)$. These thresholds are derived from the rule-based driving risk field and its corresponding safety priors \cite{wang2015driving,wang2016driving}. Given two points, if their assigned classes differ, the pairwise label is determined by the class priority. If they belong to the same class, the label is further determined by comparing their detailed risk scores:
\[
y_{ij}=
\begin{cases}
+1, & C_i > C_j; C_i=C_j,\ s_i-s_j>\epsilon,\\
-1, & C_i < C_j; C_i=C_j,\ s_j-s_i>\epsilon,\\
0,  & C_i=C_j,\ |s_i-s_j|\le \epsilon.
\end{cases}
\]
The resulting pairwise risk label set is denoted as $P_t=\{(x_i,x_j,y_{ij})\}$. This two-level design uses reliable class-level safety priors while avoiding direct regression to uncertain absolute risk scores.

\subsection{Training Losses}

DRiF is optimized with planning, auxiliary perception, static map, and dynamic risk losses:
\[
\mathcal{L}
=
\mathcal{L}_{plan}
+\lambda_{aux}\mathcal{L}_{aux}
+\lambda_{map}\mathcal{L}_{map}
+\lambda_{risk}\mathcal{L}_{risk}.
\]

\mypara{Planning and Auxiliary Losses.}
\(\mathcal{L}_{plan}\) follows the original planning objective of the base end-to-end model, such as trajectory, waypoint, target-speed, or control-related losses. \(\mathcal{L}_{aux}\) is used only in stage 1 and includes auxiliary detection, segmentation, or depth prediction losses for pretraining.

\mypara{Static Map Loss.}
The static map branch is supervised by dense BEV semantic labels. We use a standard combination of cross-entropy and Dice losses:
\[
\mathcal{L}_{map}
=
\mathcal{L}_{CE}(\hat{M}_t,M_t)
+
\mathcal{L}_{Dice}(\hat{M}_t,M_t).
\]
This loss maintains the model's understanding of static road structures during risk learning.

\mypara{Pairwise Ranking Loss.}
For each risk pair \((x_i,x_j,y_{ij})\in P_t\), we map \(x_i,x_j\) to BEV grid cells \(q_i,q_j\), and read their predicted dynamic risk values \(\hat{r}_i=\hat{R}^{dyn}_t(q_i)\) and \(\hat{r}_j=\hat{R}^{dyn}_t(q_j)\). For ordered pairs with \(y_{ij}\in\{+1,-1\}\), we use the ranking loss \cite{burges2005learning,chen2016single}:
\[
\mathcal{L}_{rank}
=
\frac{1}{|P_t|}
\sum_{(x_i,x_j,y_{ij})\in P_t}
\max\left(0,\gamma-y_{ij}(\hat{r}_i-\hat{r}_j)\right).
\]
Comparable pairs with \(y_{ij}=0\) are ignored at this stage. This objective supervises the relative ordering of risk values without regressing handcrafted absolute risk scores.


\mypara{Stage-Wise Training.}
Stage 1 pretrains the encoder with direct scene supervision:
\[
\mathcal{L}^{(1)}
=
\lambda_{aux}\mathcal{L}_{aux}.
\]
Stage 2 removes the auxiliary perception branch and jointly optimizes planning, static map understanding, and dynamic risk learning:
\[
\mathcal{L}^{(2)}
=
\mathcal{L}_{plan}
+
\lambda_{map}\mathcal{L}_{map}
+
\lambda_{risk}\mathcal{L}_{risk}.
\]

\section{Experiments}

\subsection{Experiment Settings}

\mypara{Training Dataset.}
We train DRiF on a CARLA dataset collected across multiple towns using the TF++
data collection protocol~\cite{zimmerlin2024hidden}. Each frame contains a
front-view camera image, LiDAR point cloud, ego state, route command, map
information, and expert driving trajectory. 

We generate DRiF labels from the raw logs. Static map labels are obtained
from GT BEV segmentations with dynamic objects (e.g., vehicles, walkers) removed.
For dynamic risk field labels, we sample 300 BEV candidates per frame with 50\%
strategy-guided sampling and 50\% random sampling over reachable lanes. Pairs are
ranked by staged safety comparators covering overlap risk, corridor risk and occupancy risk.

\mypara{Closed-Loop Benchmark.}
We evaluate on Bench2Drive~\cite{jia2024bench2drive}, a closed-loop CARLA
benchmark with 220 routes across diverse towns, weather, traffic, and scenarios. Given online observations and route commands, the agent
should predict future trajectories for closed-loop vehicle control.

\mypara{Evaluation Criteria.} 
Following the official Bench2Drive protocol~\cite{jia2024bench2drive}, we report Driving Score (DS), Success Rate (SR), and Driving Efficiency (Eff.); DS combines route progress and infraction penalties, while Eff. measures the route-averaged ego speed relative to surrounding traffic. We also report CRoute, the percentage of routes with at least one vehicle, pedestrian, or layout collision, and collision events per driven kilometer (Coll./km), followed by Multi-Ability (MA) scores and their mean.

\mypara{Training Configuration.}
DRiF is trained on 16 NVIDIA A100 GPUs for 60 epochs in total. 
The first 30 epochs are used for auxiliary perception pretraining, and the next 30 epochs are used for joint training with static map segmentation, dynamic risk learning, and trajectory planning objectives. 
We use an initial learning rate of \(3\times10^{-4}\), and the full training process takes about 1.5 days.

\subsection{Comparison Experiments}

\mypara{Baselines.}
We compare DRiF with representative end-to-end autonomous driving methods, including UniAD~\cite{hu2023uniad}, VAD~\cite{jiang2023vad}, ThinkTwice~\cite{jia2023thinktwice}, DriveAdapter~\cite{jia2023driveadapter}, TransFuser++  (TF++)~\cite{jaeger2023hidden}, HiP-AD~\cite{tang2025hipad}, and SimLingo~\cite{renz2025simlingo}. Among them, TF++ is a powerful baseline and achieves the best performance.

For a fair comparison, we reproduce most of the baselines from their official open-source implementations and train/evaluate them under the same data split, sensor setting, and closed-loop benchmark protocol. For UniAD and VAD, we retain the officially released benchmark scores and compute the safety statistics from their officially released route-level logs.
HiP-AD and SimLingo are not BEV-based methods and mainly operate on front-view observations, we therefore adapt their input interface and training configuration to the dataset setting while preserving their original model design. 

\mypara{Comparison Results.}
Our reproduced methods are evaluated on all 220 routes of Bench2Drive in a closed-loop manner; officially released scores are retained for UniAD and VAD.
The results are shown in Tab.~\ref{tab:bench2drive_results}.
DRiF achieves the best performance among the evaluated non-expert methods, with 88.78 DS and 75.91 SR, outperforming the strongest baseline TF++ multi-frame (4 frames) by \textbf{3.13 DS (+3.7\%)} and \textbf{6.82 SR (+9.9\%)}, while improving Eff. from 246.69 to 252.20 and reducing CRoute from 22.73\% to 15.91\%. DRiF also obtains the highest non-expert scores on most multi-ability tasks. Together, these gains demonstrate that explicit risk-field supervision improves interaction-aware planning and enables safer closed-loop driving without compromising driving efficiency.

\begin{figure*}[t]
    \centering
    \includegraphics[width=0.97\textwidth]{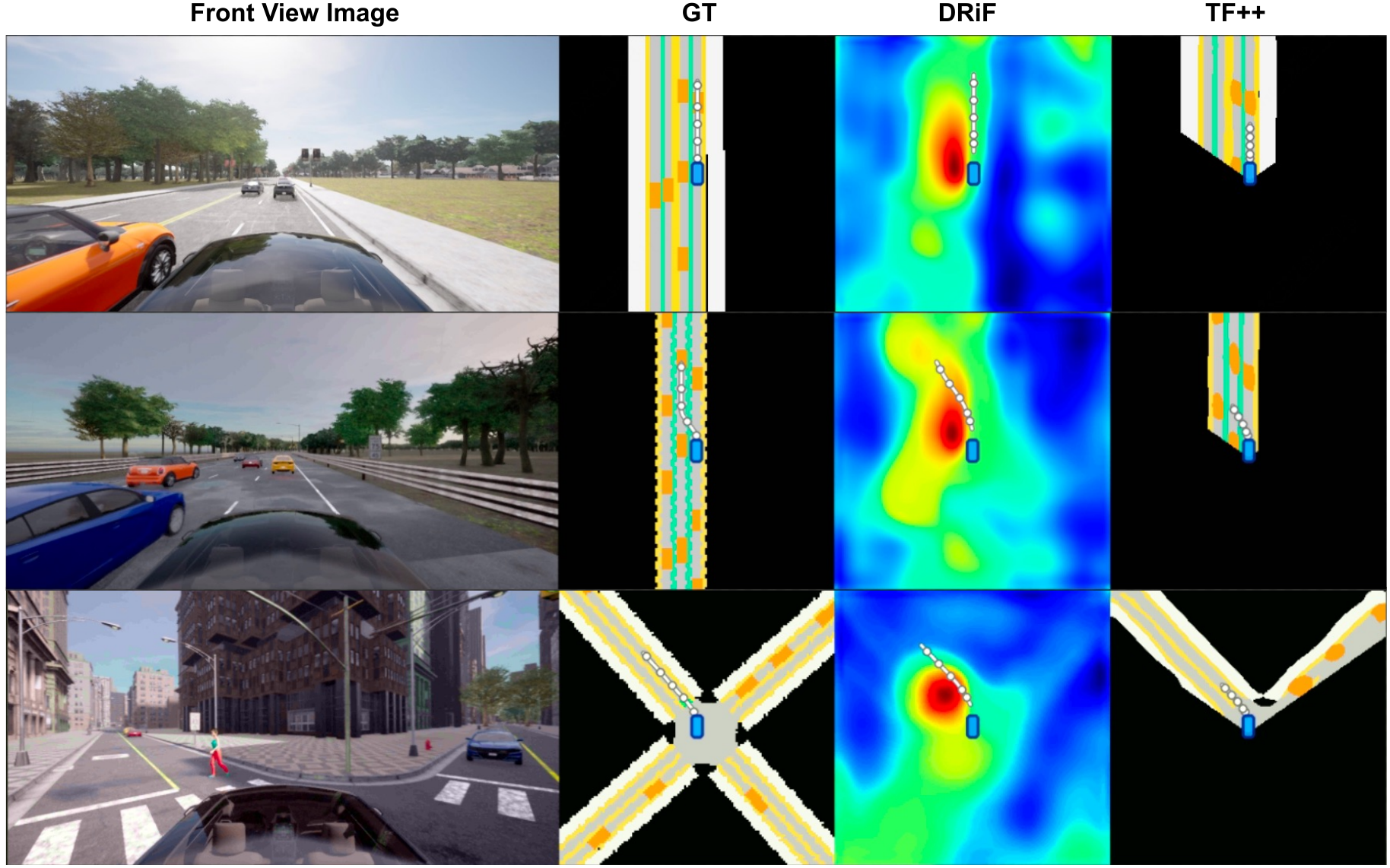}
    \caption{
  \textbf{Qualitative visualizations}. DRiF highlights ego-relevant risks and provides interpretable planning cues.
    }
    \label{fig:qualitative}
    \vspace{-1em}
\end{figure*}

\mypara{Qualitative Results.} 
As shown in Fig.~\ref{fig:qualitative}, DRiF assigns higher values to more risky areas, while keeping irrelevant occupied regions relatively low-risk.
The planned trajectories tend to avoid high-risk regions and follow safer drivable corridors, indicating that the learned risk field provides interpretable safety structure for end-to-end planning.

\begin{table*}[t]
\centering
\caption{
Closed-loop Bench2Drive results. Eff. follows the official Bench2Drive report~\cite{jia2024bench2drive}; CRoute denotes routes with at least one collision (\%), and Coll./km denotes collision events per driven kilometer. Bold and underline mark the best non-expert and baseline results, respectively. Four-frame input is used in multi-frame mode; ``$\dagger$'' marks reproduced results, ``$\ddagger$'' marks official route logs, and ``--'' denotes unavailable verifiable route records.
}
\label{tab:bench2drive_results}
\resizebox{\textwidth}{!}{
\begin{tabular}{l c c c c c c c c c c c}
\toprule
\textbf{Method}
&
\multicolumn{3}{c}{\textbf{Overall}}
&
\multicolumn{2}{c}{\textbf{Safety}}
&
\multicolumn{6}{c}{\textbf{Multi-Ability}}
\\
\cmidrule(lr){2-4}
\cmidrule(lr){5-6}
\cmidrule(lr){7-12}
&
DS $\uparrow$
&
SR $\uparrow$
&
Eff. $\uparrow$
&
CRoute $\downarrow$
&
Coll./km $\downarrow$
&
Merge $\uparrow$
&
Overtake $\uparrow$
&
EmgBrake $\uparrow$
&
GiveWay $\uparrow$
&
TSign $\uparrow$
&
Mean $\uparrow$
\\
\midrule
\textit{PDM-Lite expert}
& 97.02
& 92.27
& 250.04
& 3.64
& 0.347
& 88.75
& 93.33
& 98.33
& 90.00
& 93.68
& 92.82
\\
\midrule
UniAD~\cite{hu2023uniad}$^\ddagger$
& 45.81
& 16.36
& 129.21
& 49.08
& 9.636
& 14.10
& 17.78
& 21.67
& 10.00
& 14.21
& 15.55
\\
VAD~\cite{jiang2023vad}$^\ddagger$
& 42.35
& 15.00
& 157.94
& 42.72
& 10.674
& 8.11
& 24.44
& 18.64
& 20.00
& 19.15
& 18.07
\\
ThinkTwice~\cite{jia2023thinktwice}
& 62.44
& 31.23
& 69.33
& --
& --
& 27.38
& 18.42
& 35.82
& \textbf{\uline{50.00}}
& 54.23
& 37.17
\\
DriveAdapter~\cite{jia2023driveadapter}
& 64.22
& 33.08
& 70.22
& --
& --
& 28.82
& 26.38
& 48.76
& \textbf{\uline{50.00}}
& 56.43
& 42.08
\\
HiP-AD~\cite{tang2025hipad}$^\dagger$
& 81.88
& 62.73
& \uline{251.33}
& 26.36
& 3.528
& 53.75
& 60.00
& 73.33
& \textbf{\uline{50.00}}
& 75.26
& 62.47
\\
SimLingo~\cite{renz2025simlingo}$^\dagger$
& 84.63
& 67.73
& 250.88
& 23.18
& 2.991
& \uline{61.25}
& 53.33
& \uline{83.33}
& \textbf{\uline{50.00}}
& 84.74
& 66.53
\\
TF++~\cite{jaeger2023hidden}$^\dagger$
& 84.41
& \uline{69.09}
& 240.76
& \uline{20.45}
& \uline{2.714}
& \uline{66.25}
& 57.78
& 76.67
& \textbf{\uline{50.00}}
& 82.63
& 66.67
\\
TF++ multi-frame$^\dagger$
& \uline{85.65}
& \uline{69.09}
& 246.69
& 22.73
& 2.757
& 57.50
& \uline{64.44}
& 81.67
& \textbf{\uline{50.00}}
& \uline{85.79}
& \uline{67.88}
\\
\midrule
DRiF single-frame
& 85.21
& 67.27
& 243.86
& 25.45
& 2.861
& 58.75
& 55.56
& 78.33
& 50.00 
& 82.11
& 64.95
\\
DRiF multi-frame
& \textbf{88.78}
& \textbf{75.91}
& \textbf{252.20}
& \textbf{15.91}
& \textbf{1.955}
& \textbf{70.00}
& \textbf{68.89}
& \textbf{85.00}
& \textbf{50.00}
& \textbf{88.95}
& \textbf{72.57}
\\
\bottomrule
\end{tabular}
}
\end{table*}
\newcommand{\compacttablecaption}[2]{%
\centering
\captionof{table}{\strut #2}
\label{#1}
\vspace{-0.7em}
}
\begin{table*}[t]
\centering
\small
\setlength{\tabcolsep}{4pt}
\renewcommand{\arraystretch}{1.08}

\begin{minipage}[t]{0.49\textwidth}
\compacttablecaption{ablation:1}{Static map and dynamic risk ablations.}
\begin{tabular*}{\linewidth}{@{\hspace{0.05\linewidth}}l@{\extracolsep{\fill}}ccc@{\hspace{0.05\linewidth}}}
\toprule
Setting & DS & SR & MA-Mean \\
\midrule
w/o Static & 83.23 & 60.00 & 63.97 \\
w/o Risk & 62.40 & 45.45 & 35.03 \\
DRiF & \textbf{88.78} & \textbf{75.91} & \textbf{72.57} \\
\bottomrule
\end{tabular*}
\end{minipage}
\hfill
\begin{minipage}[t]{0.49\textwidth}
\compacttablecaption{ablation:4}{Risk sources ablations.}
\begin{tabular*}{\linewidth}{@{\hspace{0.05\linewidth}}l@{\extracolsep{\fill}}ccc@{\hspace{0.05\linewidth}}}
\toprule
Sources & DS & SR & MA-Mean \\
\midrule
\(R_3\) (Occupancy) & 86.29 & 70.45 & 65.90 \\
\(R_1 + R_2\) & 87.77 & 72.73 & 69.28 \\
\(R_1 + R_2 + R_3\) & \textbf{88.78} & \textbf{75.91} & \textbf{72.57} \\
\bottomrule
\end{tabular*}
\end{minipage}

\vspace{0.8em}

\begin{minipage}[t]{0.49\textwidth}
\compacttablecaption{ablation:2}{Absolute value supervision ablation.}
\begin{tabular*}{\linewidth}{@{\hspace{0.035\linewidth}}l@{\extracolsep{\fill}}ccc@{\hspace{0.035\linewidth}}}
\toprule
Setting & DS & SR & MA-Mean \\
\midrule
Absolute & 85.41 & 70.45 & 65.38 \\
Relative & \textbf{88.78} & \textbf{75.91} & \textbf{72.57} \\
\bottomrule
\end{tabular*}
\end{minipage}
\hfill
\begin{minipage}[t]{0.49\textwidth}
\compacttablecaption{ablation:5}{Auxiliary task ablation.}
\begin{tabular*}{\linewidth}
{@{\hspace{0.035\linewidth}}l@{\extracolsep{\fill}}ccc@{\hspace{0.035\linewidth}}}
\toprule
Setting & DS & SR & MA-Mean \\
\midrule
w/ Stage-2 Aux & \textbf{88.88} & 74.55 & 70.02 \\
w/o Stage-2 Aux & 88.78 & \textbf{75.91} & \textbf{72.57} \\
\bottomrule
\end{tabular*}
\end{minipage}
\vspace{-1em}
\end{table*}

\subsection{Ablation Studies}

All ablation studies use 4-frame inputs and start from the same baseline
stage-1 checkpoint. The default DRiF setting uses relative risk supervision from
all three sources \(R_1+R_2+R_3\), and static map semantics supervision.

\mypara{Static Map and Risk Supervision.}
We ablate the two main supervision signals in DRiF. Removing the static map tests the
effect of static road-context supervision, while removing the risk branch tests
the contribution of interaction-aware relative risk learning. The results are listed in Tab. \ref{ablation:1}. The results show
that both branches are beneficial, indicating that static road-context cues and
dynamic risk cues provide complementary information for closed-loop
driving.

\mypara{Risk Sources.}
We ablate the risk sources used for pair ranking. \(R_1\), \(R_2\), and \(R_3\)
denote overlap risk, corridor risk, and occupancy risk, respectively. We compare variants using \(R_3\) (i.e., occupancy),
\(R_1+R_2\), and \(R_1+R_2+R_3\) (i.e., DRiF). The results in Tab. \ref{ablation:4} show that combining
all three sources achieves the best performance, indicating the necessity of all risk classes.

\mypara{Absolute Value Supervision.}
We replace pairwise relative supervision with absolute score regression. This
examines whether learning risk ordering between BEV locations is more effective
than directly fitting handcrafted scalar risk scores. The results are shown in Tab. \ref{ablation:2}. Relative supervision outperforms absolute-score regression by 3.37 DS, 5.46 SR, and 7.19 MA-Mean, supporting the advantage of learning risk ordering without fitting manually calibrated magnitudes.


\mypara{Auxiliary Tasks.}
Finally, we add the original auxiliary perception tasks in the 2nd training stage, including
semantic, depth, and detection supervision. This tests whether DRiF can perceive and understand the surroundings only based on risk-oriented training
objective. Based on the results in Tab. \ref{ablation:5}, we find that the proposed risk-oriented supervision
already provides compact and task-relevant scene understanding for driving.

\subsection{Limitations}

Currently, the pairwise label generator only covers several representative and easy-to-validate risk sources (i.e., overlap, corridor and occupancy risks). This simplified design is mainly for quick proof-of-concept validation in CARLA. Thus, the current risk-pair labels do not yet reflect a fully comprehensive risk field for real-world applications. Future work will use more complete risk theories to guide the data-driven risk-field learning, especially on large-scale real-vehicle data.

\section{Conclusion}

We presented \textbf{DRiF}, a data-driven risk-field framework that learns a shared BEV representation for static-map understanding, dynamic risk prediction, and trajectory planning. Pairwise risk labels supervise relative ordering instead of handcrafted absolute values, retaining interpretable safety structure while enabling data-driven learning. Closed-loop Bench2Drive results show improvements in driving score, success rate, and interaction abilities, supporting explicit risk-field supervision for end-to-end planning.

\bibliographystyle{IEEEtran}
\bibliography{references}

@article{bojarski2016end,
  title={End to End Learning for Self-Driving Cars},
  author={Bojarski, Mariusz and {Del Testa}, Davide and Dworakowski, Daniel and Firner, Bernhard and Flepp, Beat and Goyal, Prasoon and Jackel, Lawrence D. and Monfort, Mathew and Muller, Urs and Zhang, Jiakai and Zhang, Xin and Zhao, Jake and Zieba, Karol},
  journal={arXiv preprint arXiv:1604.07316},
  year={2016}
}

@inproceedings{bansal2019chauffeurnet,
  title={{ChauffeurNet}: Learning to Drive by Imitating the Best and Synthesizing the Worst},
  author={Bansal, Mayank and Krizhevsky, Alex and Ogale, Abhijit S.},
  booktitle={RSS},
  year={2019}
}

@article{chitta2023transfuser,
  title={{TransFuser}: Imitation with Transformer-Based Sensor Fusion for Autonomous Driving},
  author={Chitta, Kashyap and Prakash, Aditya and Jaeger, Bernhard and Yu, Zehao and Renz, Katrin and Geiger, Andreas},
  journal={IEEE TPAMI},
  year={2023}
}

@article{chen2024end,
  title={End-to-End Autonomous Driving: Challenges and Frontiers},
  author={Chen, Li and Wu, Penghao and Chitta, Kashyap and Jaeger, Bernhard and Geiger, Andreas and Li, Hongyang},
  journal={IEEE TPAMI},
  year={2024}
}

@inproceedings{philion2020lift,
  title={Lift, Splat, Shoot: Encoding Images from Arbitrary Camera Rigs by Implicitly Unprojecting to 3D},
  author={Philion, Jonah and Fidler, Sanja},
  booktitle={ECCV},
  year={2020}
}

@inproceedings{li2022bevformer,
  title={{BEVFormer}: Learning Bird's-Eye-View Representation from Multi-Camera Images via Spatiotemporal Transformers},
  author={Li, Zhiqi and Wang, Wenhai and Li, Hongyang and Xie, Enze and Sima, Chonghao and Lu, Tong and Yu, Qiao and Dai, Jifeng},
  booktitle={ECCV},
  year={2022}
}

@inproceedings{liu2022bevfusion,
  title={{BEVFusion}: Multi-Task Multi-Sensor Fusion with Unified Bird's-Eye View Representation},
  author={Liu, Zhijian and Tang, Haotian and Amini, Alexander and Yang, Xinyu and Mao, Huizi and Rus, Daniela and Han, Song},
  booktitle={ICRA},
  year={2023}
}

@inproceedings{liao2022maptr,
  title={{MapTR}: Structured Modeling and Learning for Online Vectorized {HD} Map Construction},
  author={Liao, Bencheng and Chen, Shaoyu and Wang, Xinggang and Cheng, Tianheng and Zhang, Qian and Liu, Wenyu and Huang, Chang},
  booktitle={ICLR},
  year={2023}
}

@inproceedings{jiang2023vad,
  title={{VAD}: Vectorized Scene Representation for Efficient Autonomous Driving},
  author={Jiang, Bo and Chen, Shaoyu and Xu, Qing and Liao, Bencheng and Chen, Jiajie and Zhou, Helong and Zhang, Qian and Liu, Wenyu and Huang, Chang and Wang, Xinggang},
  booktitle={ICCV},
  year={2023}
}

@inproceedings{hu2021fiery,
  title={{FIERY}: Future Instance Prediction in Bird's-Eye View from Surround Monocular Cameras},
  author={Hu, Anthony and Murez, Zak and Mohan, Nikhil and Dudas, Sof{\'i}a and Hawke, Jeffrey and Badrinarayanan, Vijay and Cipolla, Roberto and Kendall, Alex},
  booktitle={ICCV},
  year={2021}
}

@inproceedings{shi2022motion,
  title={Motion Transformer with Global Intention Localization and Local Movement Refinement},
  author={Shi, Shaoshuai and Jiang, Li and Dai, Dengxin and Schiele, Bernt},
  booktitle={NeurIPS},
  year={2022}
}

@inproceedings{wei2023surroundocc,
  title={{SurroundOcc}: Multi-Camera 3D Occupancy Prediction for Autonomous Driving},
  author={Wei, Yi and Zhao, Linqing and Zheng, Wenzhao and Zhu, Zheng and Zhou, Jie and Lu, Jiwen},
  booktitle={ICCV},
  year={2023}
}

@inproceedings{zhang2023occformer,
  title={{OccFormer}: Dual-Path Transformer for Vision-Based 3D Semantic Occupancy Prediction},
  author={Zhang, Yunpeng and Zhu, Zheng and Du, Dalong},
  booktitle={ICCV},
  year={2023}
}

@inproceedings{hu2022stp3,
  title={{ST-P3}: End-to-End Vision-Based Autonomous Driving via Spatial-Temporal Feature Learning},
  author={Hu, Shengchao and Chen, Li and Wu, Penghao and Li, Hongyang and Yan, Junchi and Tao, Dacheng},
  booktitle={ECCV},
  year={2022}
}

@inproceedings{hu2023uniad,
  title={Planning-Oriented Autonomous Driving},
  author={Hu, Yihan and Yang, Jiazhi and Chen, Li and Li, Keyu and Sima, Chonghao and Zhu, Xizhou and Chai, Siqi and Du, Senyao and Lin, Tianwei and Wang, Wenhai and Lu, Lewei and Jia, Xiaosong and Liu, Qiang and Dai, Jifeng and Qiao, Yu and Li, Hongyang},
  booktitle={CVPR},
  year={2023}
}

@inproceedings{sun2024sparsedrive,
  title={{SparseDrive}: End-to-End Autonomous Driving via Sparse Scene Representation},
  author={Sun, Wenchao and Lin, Xuewu and Shi, Yining and Zhang, Chuang and Wu, Haoran and Zheng, Sifa},
  booktitle={ICRA},
  year={2025}
}

@inproceedings{song2025momad,
  title={Don't Shake the Wheel: Momentum-Aware Planning in End-to-End Autonomous Driving},
  author={Song, Ziying and Jia, Caiyan and Liu, Lin and Pan, Hongyu and Zhang, Yongchang and Wang, Junming and Zhang, Xingyu and Xu, Shaoqing and Yang, Lei and Luo, Yadan},
  booktitle={CVPR},
  year={2025}
}

@inproceedings{liao2025diffusiondrive,
  title={{DiffusionDrive}: Truncated Diffusion Model for End-to-End Autonomous Driving},
  author={Liao, Bencheng and Chen, Shaoyu and Yin, Haoran and Jiang, Bo and Wang, Cheng and Yan, Sixu and Zhang, Xinbang and Li, Xiangyu and Zhang, Ying and Zhang, Qian and Wang, Xinggang},
  booktitle={CVPR},
  year={2025}
}

@inproceedings{xing2025goalflow,
  title={{GoalFlow}: Goal-Driven Flow Matching for Multimodal Trajectories Generation in End-to-End Autonomous Driving},
  author={Xing, Zebin and Zhang, Xingyu and Hu, Yang and Jiang, Bo and He, Tong and Zhang, Qian and Long, Xiaoxiao and Yin, Wei},
  booktitle={CVPR},
  year={2025}
}

@inproceedings{li2025wote,
  title={End-to-End Driving with Online Trajectory Evaluation via {BEV} World Model},
  author={Li, Yingyan and Wang, Yuqi and Liu, Yang and He, Jiawei and Fan, Lue and Zhang, Zhaoxiang},
  booktitle={ICCV},
  year={2025}
}

@inproceedings{zheng2025world4drive,
  title={{World4Drive}: End-to-End Autonomous Driving via Intention-aware Physical Latent World Model},
  author={Zheng, Yupeng and Yang, Pengxuan and Xing, Zebin and Zhang, Qichao and Zheng, Yuhang and Gao, Yinfeng and Li, Pengfei and Zhang, Teng and Xia, Zhongpu and Jia, Peng and Lang, XianPeng and Zhao, Dongbin},
  booktitle={ICCV},
  year={2025}
}

@inproceedings{zhang2025seerdrive,
  title={Future-Aware End-to-End Driving: Bidirectional Modeling of Trajectory Planning and Scene Evolution},
  author={Zhang, Bozhou and Song, Nan and Li, Jingyu and Zhu, Xiatian and Deng, Jiankang and Zhang, Li},
  booktitle={NeurIPS},
  year={2025}
}

@article{hayward1972near,
  title={Near-Miss Determination Through Use of a Scale of Danger},
  author={Hayward, John C.},
  journal={Highway Research Record},
  year={1972}
}

@article{zhang2026resworld,
  title={{ResWorld}: Temporal Residual World Model for End-to-End Autonomous Driving},
  author={Zhang, Jinqing and Fu, Zehua and Xu, Zelin and Dai, Wenying and Liu, Qingjie and Wang, Yunhong},
  journal={arXiv preprint arXiv:2602.10884},
  year={2026}
}

@article{lei2025riskmm,
  title={Risk Map as Middleware: Towards Interpretable Cooperative End-to-End Autonomous Driving for Risk-Aware Planning},
  author={Lei, Mingyue and Zhou, Zewei and Li, Hongchen and Ma, Jiaqi and Hu, Jia},
  journal={IEEE RA-L},
  year={2026}
}

@article{vogel2003comparison,
  title={A Comparison of Headway and Time to Collision as Safety Indicators},
  author={Vogel, Katja},
  journal={Accident Analysis \& Prevention},
  year={2003}
}

@article{cooper1976traffic,
  title={Traffic Studies at {T}-Junctions. 2. A Conflict Simulation Record},
  author={Cooper, D. F. and Ferguson, N.},
  journal={Traffic Engineering \& Control},
  year={1976}
}

@incollection{cooper1984experience,
  title={Experience with Traffic Conflicts in Canada with Emphasis on ``Post Encroachment Time'' Techniques},
  author={Cooper, Peter J.},
  booktitle={International Calibration Study of Traffic Conflict Techniques},
  year={1984}
}

@article{fu2021comparison,
  title={Comparison of Threshold Determination Methods for the Deceleration Rate to Avoid a Crash ({DRAC})-Based Crash Estimation},
  author={Fu, Chuanyun and Sayed, Tarek},
  journal={Accident Analysis \& Prevention},
  year={2021}
}

@article{shalev2017formal,
  title={On a Formal Model of Safe and Scalable Self-Driving Cars},
  author={Shalev-Shwartz, Shai and Shammah, Shaked and Shashua, Amnon},
  journal={arXiv preprint arXiv:1708.06374},
  year={2017}
}

@article{khatib1986real,
  title={Real-Time Obstacle Avoidance for Manipulators and Mobile Robots},
  author={Khatib, Oussama},
  journal={IJRR},
  year={1986}
}

@article{wang2015driving,
  title={The Driving Safety Field Based on Driver--Vehicle--Road Interactions},
  author={Wang, Jianqiang and Wu, Jian and Li, Yang},
  journal={IEEE T-ITS},
  year={2015}
}

@article{wang2016driving,
  title={Driving Safety Field Theory Modeling and Its Application in Pre-Collision Warning System},
  author={Wang, Jianqiang and Wu, Jian and Zheng, Xunjia and Ni, Daiheng and Li, Keqiang},
  journal={TR-C},
  year={2016}
}

@article{mullakkal2020probabilistic,
  title={Probabilistic Field Approach for Motorway Driving Risk Assessment},
  author={{Mullakkal-Babu}, Freddy A. and Wang, Meng and He, Xiaolin and {van Arem}, Bart and Happee, Riender},
  journal={TR-C},
  year={2020}
}

@article{wang2022probabilistic,
  title={Probabilistic Risk Metric for Highway Driving Leveraging Multi-Modal Trajectory Predictions},
  author={Wang, Xinwei and Alonso-Mora, Javier and Wang, Meng},
  journal={IEEE T-ITS},
  year={2022}
}

@article{wang2026data,
  title={A Data-Driven Spatio-Temporal Driving Risk Field Mechanism for Path Planning},
  author={Wang, Zhuoer and Shi, Baohan and Zhang, Jianping and Zhu, Xiaowen and Zhou, Jian and Xu, Bingrong and Li, Bijun},
  journal={Expert Systems with Applications},
  year={2026}
}

@article{bradley1952rank,
  title={Rank Analysis of Incomplete Block Designs: I. The Method of Paired Comparisons},
  author={Bradley, Ralph Allan and Terry, Milton E.},
  journal={Biometrika},
  year={1952}
}

@inproceedings{burges2005learning,
  title={Learning to Rank Using Gradient Descent},
  author={Burges, Christopher J. C. and Shaked, Tal and Renshaw, Erin and Lazier, Ari and Deeds, Matt and Hamilton, Nicole and Hullender, Gregory N.},
  booktitle={ICML},
  year={2005}
}

@inproceedings{xin2024riskmap,
  title={{RiskMap}: A Unified Driving Context Representation for Autonomous Motion Planning in Urban Driving Environment},
  author={Xin, Ren and Wang, Sheng and Chen, Yingbing and Cheng, Jie and Liu, Ming and Ma, Jun},
  booktitle={ROBIO},
  year={2024}
}

@article{chen2024riskoccupancy,
  title={Risk Occupancy: A New and Efficient Paradigm through Vehicle-Road-Cloud Collaboration},
  author={Chen, Jiaxing and Zhong, Wei and Gao, Bolin and Liu, Yifei and Zou, Hengduo and Liu, Jiaxi and Lu, Yanbo and Huang, Jin and Zhong, Zhihua},
  journal={arXiv preprint arXiv:2408.07367},
  year={2024}
}

@inproceedings{chen2016single,
  title={Single-Image Depth Perception in the Wild},
  author={Chen, Weifeng and Fu, Zhao and Yang, Dawei and Deng, Jia},
  booktitle={NeurIPS},
  year={2016}
}

@inproceedings{fu2018deep,
  title={Deep Ordinal Regression Network for Monocular Depth Estimation},
  author={Fu, Huan and Gong, Mingming and Wang, Chaohui and Batmanghelich, Kayhan and Tao, Dacheng},
  booktitle={CVPR},
  year={2018}
}

@inproceedings{yang2024depthanythingv2,
  title={Depth Anything V2},
  author={Yang, Lihe and Kang, Bingyi and Huang, Zilong and Zhao, Zhen and Xu, Xiaogang and Feng, Jiashi and Zhao, Hengshuang},
  booktitle={NeurIPS},
  year={2024}
}

@article{xu2024drivegpt4,
  title={{DriveGPT4}: Interpretable End-to-End Autonomous Driving via Large Language Model},
  author={Xu, Zhenhua and Zhang, Yujia and Xie, Enze and Zhao, Zhen and Guo, Yong and Wong, Kwan-Yee K. and Li, Zhenguo and Zhao, Hengshuang},
  journal={IEEE RA-L},
  year={2024}
}

@inproceedings{xu2025drivegpt4v2,
  title={{DriveGPT4-V2}: Harnessing Large Language Model Capabilities for Enhanced Closed-Loop Autonomous Driving},
  author={Xu, Zhenhua and Bai, Yan and Zhang, Yujia and Li, Zhuoling and Xia, Fei and Wong, Kwan-Yee K. and Wang, Jianqiang and Zhao, Hengshuang},
  booktitle={CVPR},
  year={2025}
}

@inproceedings{xu2024insmapper,
  title={{InsMapper}: Exploring Inner-Instance Information for Vectorized {HD} Mapping},
  author={Xu, Zhenhua and Wong, Kwan-Yee K. and Zhao, Hengshuang},
  booktitle={ECCV},
  year={2024}
}

@article{zimmerlin2024hidden,
  title={Hidden Biases of End-to-End Driving Datasets},
  author={Zimmerlin, Julian and Bei{\ss}wenger, Jens and Jaeger, Bernhard and Geiger, Andreas and Chitta, Kashyap},
  journal={arXiv preprint arXiv:2412.09602},
  year={2024}
}

@inproceedings{jia2024bench2drive,
  title={Bench2Drive: Towards Multi-Ability Benchmarking of Closed-Loop End-To-End Autonomous Driving},
  author={Jia, Xiaosong and others},
  booktitle={NeurIPS},
  year={2024}
}

@inproceedings{jia2023thinktwice,
  title={Think Twice before Driving: Towards Scalable Decoders for End-to-End Autonomous Driving},
  author={Jia, Xiaosong and Wu, Penghao and Chen, Li and Xie, Jiangwei and He, Conghui and Yan, Junchi and Li, Hongyang},
  booktitle={CVPR},
  year={2023}
}

@inproceedings{jia2023driveadapter,
  title={DriveAdapter: Breaking the Coupling Barrier of Perception and Planning in End-to-End Autonomous Driving},
  author={Jia, Xiaosong and Gao, Yulu and Chen, Li and Yan, Junchi and Liu, Patrick Langechuan and Li, Hongyang},
  booktitle={ICCV},
  year={2023}
}

@inproceedings{jaeger2023hidden,
  title={Hidden Biases of End-to-End Driving Models},
  author={Jaeger, Bernhard and Chitta, Kashyap and Geiger, Andreas},
  booktitle={ICCV},
  year={2023}
}

@inproceedings{renz2025simlingo,
  title={SimLingo: Vision-Only Closed-Loop Autonomous Driving with Language-Action Alignment},
  author={Renz, Katrin and Chen, Long and Arani, Elahe and Sinavski, Oleg},
  booktitle={CVPR},
  year={2025}
}

@inproceedings{tang2025hipad,
  title={HiP-AD: Hierarchical and Multi-Granularity Planning with Deformable Attention for Autonomous Driving in a Single Decoder},
  author={Tang, Yingqi and Xu, Zhuoran and Meng, Zhaotie and Cheng, Erkang},
  booktitle={ICCV},
  year={2025}
}

@article{moon2026caad,
  title={Causality-Aware End-to-End Autonomous Driving via Ego-Centric Joint Scene Modeling},
  author={Moon, Seokha and Lee, Minseung and Seo, Joon and Kim, Jinkyu and Lee, Jungbeom},
  journal={arXiv preprint arXiv:2605.13646},
  year={2026}
}

@article{cheng2026owmdrive,
  title={OWMDrive: Causality-Aware End-to-End Autonomous Driving via 4D Occupancy World Model},
  author={Cheng, Junjie and Song, Ruiqi and Wu, Ye and Zeng, Nanxing and Li, Ximiao and Ai, Yunfeng},
  journal={arXiv preprint arXiv:2606.30421},
  year={2026}
}

@article{wu2026aligndrive,
  title={AlignDrive: Aligned Lateral-Longitudinal Planning for End-to-End Autonomous Driving},
  author={Wu, Yanhao and Zhang, Haoyang and He, Fei and Wu, Rui and Qiu, Congpei and Gao, Liang and Ke, Wei and Zhang, Tong},
  journal={arXiv preprint arXiv:2601.01762},
  year={2026}
}

@IEEEtranBSTCTL{IEEEtran:BSTcontrol,
  CTLuse_forced_etal       = "yes",
  CTLmax_names_forced_etal = "6",
  CTLnames_show_etal       = "1",
  CTLdash_repeated_names    = "no"
}

\end{document}